# Accelerating Deep Learning Model Inference on Arm CPUs with Ultra-Low Bit Quantization and Runtime

Saad Ashfaq, MohammadHossein AskariHemmat, Sudhakar Sah, Ehsan Saboori, Olivier Mastropietro, Alexander Hoffman
Deeplite Inc.
Montreal, Canada
(saad, hossein, sudhakar, ehsan, olivier, alexander.hoffman) @deeplite.ai

*Abstract*—Deep Learning has been one of the most disruptive technological advancements in recent times. The high performance of deep learning models comes at the expense of high computational, storage and power requirements. Sensing the immediate need for accelerating and compressing these models to improve on-device performance, we introduce Deeplite Neutrino for production-ready optimization of the models and Deeplite Runtime for deployment of ultra-low bit quantized models on Arm-based platforms. We implement low-level quantization kernels for Armv7 and Armv8 architectures enabling deployment on the vast array of 32-bit and 64-bit Arm-based devices. With efficient implementations using vectorization, parallelization, and tiling, we realize speedups of up to 2x and 2.2x compared to TensorFlow Lite with XNNPACK backend on classification and detection models, respectively. We also achieve significant speedups of up to 5x and 3.2x compared to ONNX Runtime for classification and detection models, respectively.

## I. INTRODUCTION

Modern deep learning models for computer vision are achieving accuracies and detection performance with the potential to drive exciting new product developments across multiple domains, including but not limited to, home surveillance cameras, AI traffic monitoring systems, automotive driver assistance, mobile video applications and intelligent consumer electronics. It is becoming increasingly practical to deploy deep learning models for vision applications on existing server infrastructure, especially with the advent of high-performance data center GPUs and CPUs. However, due to stringent requirements for deterministic latency, reliable connectivity, data privacy, and cost sensitivity of many applications, the use of cloud-centric AI has become a major limiting factor in many vision applications. For example, there are privacy issues related to uploading non-anonymized facial data in face detection for surveillance, real-time person detection in automotive requires on device processing and always-on person ID with smart doorbell cameras needs cost-effective inference performance. Therefore, to achieve mass-market deployment, cutting-edge vision applications need to be enabled within edge devices themselves. The lack of sufficient processing power and practical economics to run these models in real-time, all the time, has been a major barrier to adoption.

Model quantization is one of the ways to improve the performance of computer vision models on CPUs and overcome barriers for adopting AI at the edge. Current quantization frameworks and inference engines predominantly use 8-bit integer (INT8) quantization for model weights and activations, instead of using the same precision as training, typically 32-bit floating-point (FP32). This reduces memory bandwidth and can potentially provide 2-4x speedup while retaining accuracy. However, we sought to push the limits of existing quantization methods by using less than 8-bits for quantization.

In this paper, we present Deeplite Runtime (DeepliteRT), an innovative ultra-low precision inference engine for Arm CPUs, and a quantization framework in Deeplite Neutrino™ [4] that automatically quantizes and optimizes CNN models with less than 4 bits of precision. DeepliteRT empowers developers to unlock advanced AI on billions of low-power Arm-based devices achieving up to 5x faster inference compared to optimized FP32 baselines and up to 2x faster inference for classification models relative to TFLite [10] used with the highly optimized XNNPack [12] backend. The speedups on object detection models are up to 3.2x and 2.2x compared to ONNX Runtime and TFLite with XNNpack, respectively.

## II. BACKGROUND

Deep Learning model performance has progressed considerably in the past few years, especially with state-of-the-art classification [3] and object detection models [1], [17]. These models perform well on computer vision problems and improve upon the state-of-the-art year-over-year. However, constantly improving performance of these models still does not make them suitable to be used on edge devices due to computational, power and memory requirements. These models use 32-bit precision which can be reduced to 16 bits or even to 8 bits by quantizing the weights and activations accordingly.

Existing methods for neural network quantization can be classified in two categories [24], Post-training Quantization (PTQ) and Quantization-Aware Training (QAT). PTQ does not require the original network to be re-trained on the target dataset and is therefore typically limited to FP16 or INT8 quantization, where sufficient information is preserved in the original training statistics to accurately quantize the network. On the other hand, QAT quantizes the network during training and preserves significantly more accuracy than PTQ. QAT methods have been tested at as low as INT4 precision [23] using 4 bits per neural network weight and 8 bits per activation. However, INT4 quantization in [23] is tested only on classification models. Low bit quantization is much more difficult in object detection problems. Significant work to quantize the model using just 1 bit is also available but such models suffer from accuracy loss beyond 10% [26][27].

On the other end of the research spectrum, low complexity light weight models like MobileNet [2] are available to help enable AI on edge devices. However, these models are handcrafted and lack the performance needed for many computer vision applications. Also, because of their handcrafted and compact architecture, compression techniques such as quantization are generally not effective [18] and custom modifications might be necessary to accommodate quantization [22].

Even when high accuracy quantized models with 1-2 bits of precision are available, they cannot be deployed on commercial off-the-shelf hardware [19] due to lack of support for 1 or 2-bit instructions and operators. Most commodity hardware only supports down to 8-bit operations. Such devices either need a custom inference engine (runtime) or custom hardware support like INT4 instructions provided in advanced NVIDIA processors [25]. Therefore, ultra-low bit quantization (below 4-bits) remains a challenging problem to solve both from the QAT as well as the runtime perspective. Currently, standard AI frameworks such as TensorFlow [15] and PyTorch [16] do not support 1 bit or 2 bits of precision. Open-source runtimes and compilers such as TFLite [10] and TVM [28] currently support 8 bits of precision with quantization.

This paper makes the following contributions:
- Quantization aware training with 1 and 2-bit precision (weights and activations) for object detection and classification models.
- Mixed precision approach to minimize the accuracy drop of quantized models.
- Custom ultra-low precision convolution operators to accelerate speed and memory throughput of quantized layers.
- End-to-end framework to deploy and execute mixed precision ultra-low bit quantized models on Armv7 and Armv8 Cortex-A processors.

### III. CHALLENGES OF DETECTION MODELS FOR EMBEDDED APPLICATIONS

We benchmarked state-of-the-art object detection models on a Raspberry Pi 4B platform. Our initial study found that unless we use a very compact model (YOLOv5n [1]) combined with a very low-resolution input image (less than 300px), we cannot achieve more than 4-5 FPS, even with 8-bit quantization (Figure 1). The speedup from using YOLOv5n, however, has an expensive tradeoff with accuracy as shown in the graph in Figure 2.

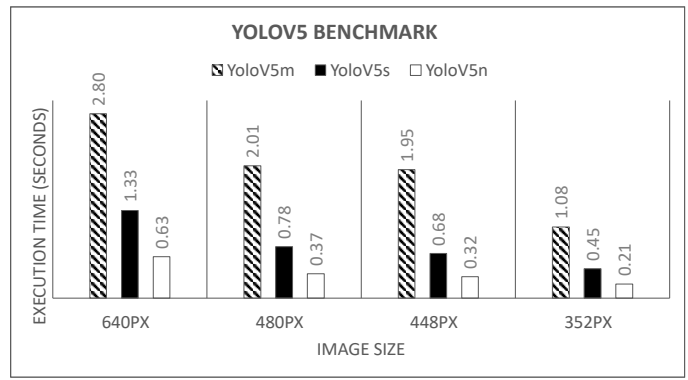

Fig. 1.  YOLOv5 benchmark on Raspberry Pi 4B (Arm Cortex A-72)

The high latency and low throughput for current deep neural networks on commodity CPUs like the Cortex-A72 in the Raspberry Pi 4B demonstrates the harsh limitations of AI inference on low power and affordable processors. Despite there being billions of devices powered by ARM Cortex-A CPUs, even the latest quantization techniques did not provide sufficiently low latency numbers for practical applications. Although reducing the model parameters from 32 bits to 8 bits results in respectable speedup without a significant loss in accuracy, it is still not enough to run these models on such small footprint hardware. Furthermore, many of the compact networks [2] designed for these devices were not accurate enough, including the smaller variations of YOLOv5n [1] as shown in Figure 2.

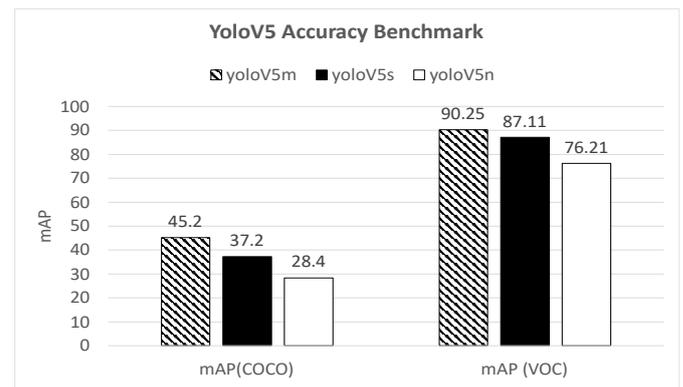

Fig. 2.  Accuracy drop on YOLO Variants for VOC and COCO datasets [1]

### IV. ACCURATE ULTRA-LOW BIT QUANTIZATION

It has been known for several years now that deep learning models are robust to quantization. However, how many bits is enough for certain AI tasks remains an open-ended question ever since early work on post-training and training-aware quantization, such as in [5] and [6].

Using advances in Deeplite Neutrino™, a CNN based model optimization software, we employed a quantization-aware training method to quantize both model weights and activations below 4 bits. In addition, we provide a way to save and infer such models on low-power CPUs and compare to full-precision and INT8 baselines using DeepliteRT.

The main goal of quantization is to reduce model precision and complexity while preserving accuracy. As mentioned earlier, quantization can be applied to a model after or while training. Quantization after training (post-training quantization) can be done statically or dynamically. In post-training static quantization, weights are quantized ahead of time, using a calibration process on the validation set to compute a scale and bias for the activations. In post-training dynamic quantization, much like post-training static quantization, the weights are quantized ahead of time, but the activations are dynamically quantized at inference. Dynamic quantization is useful for models where model execution time is dominated by the time it takes to load weights for the model e.g., LSTM [30]. Quantization can also be learned by the network. In quantization-aware training [6], the model will learn to represent parameters with a pre-defined precision. Compared to post-training quantization, quantization-aware training yields better accuracy. On the other hand, it takes another training session to learn the newly quantized parameters.

Neutrino provides state-of-the-art quantization aware techniques that can quantize models down to 3 bits, 2 bits or 1 bit as well as mixed precision. At training, we quantize input tensor t as below [4]:

$$\bar{t} = \left| \text{clip}\left(\frac{t}{s}, -Q_N, Q_P\right) \right|$$

Where, $\bar{t} \in \mathbb{N}$ is the quantized tensor, $t \in \mathbb{R}$ is the input tensor and $s \in \mathbb{R}$ is the scaling factor. To quantize the input tensor t with b bits, $Q_P = 2^{b-1} - 1$ and $Q_N = 2^{b-1}$ represent the clipping limits. Given the equation above, the quantization error can be computed as below:

$$\hat{t} = \bar{t} \times s, \quad \text{error}_q = t - \hat{t}$$

At training, our quantization algorithm learns the scaling factor so that the quantization error, $\text{error}_q$, will be minimum. Our quantizer is coupled with DeepliteRT to efficiently run the quantized model on the target platform.

## V. DeepliteRT

To enable extremely low precision computations on commodity hardware such as Arm CPUs, we require a proprietary inference engine. Public domain AI runtimes, such as TFLite and ONNX runtime [11] currently only support quantization precision down to 8-bit. Furthermore, running operations in less than 8-bit either requires special hardware support or innovative ways to perform low bit computations. Running models with sub-8 bit precision on commodity hardware has been explored before. TernGEMM [32] proposes a GEMM based method that uses logical bitwise operators to perform matrix multiplication for ternary weights, {-1, 0, 1}, and 3 to 6-bit activations. Han et al. [13], developed low precision convolution kernels and optimization passes in the TVM machine learning compiler to compute sub- 8-bit operations on commodity hardware. ULLPACK [33] presents two packing schemes for low precision operations that allow for a trade-off between accuracy and speed.

DeepliteRT leverages specific low-level hardware intrinsic to run our ultra-low precision quantized models with high performance on Arm CPUs. DeepliteRT accelerates model inference through ultra-low bit convolution kernels. The convolution is performed using bitserial computation where popcount and bitwise operations are utilized to calculate the dot products of the low bit weight and activation values. Using 1-bit weights and activations with unipolar encoding where each bit can take on the values {0, 1}, the bitserial dot product can be computed with the following equation:

$$W \cdot A = POPCOUNT\ (W\ \&\ A)$$

This can be further extended to the multi-bit case (w bits for weights and a bits for activations) by splitting the weights and activations into separate bitplanes and then summing the dot products across all the bitplane combinations:

$$W \cdot A = \sum_{i=0}^{w-1} \sum_{j=0}^{a-1} \left(POPCOUNT\ (W[i]\ \&\ A[j]) \ll (i+j)\right)$$

Our implementation uses intrinsic from the Neon vectorized instruction set for both Armv7 and Armv8 architectures to target 32-bit and 64-bit Arm CPU devices. Efficient tiling and parallelization schemes are also used to improve upon the performance of the vectorized kernels. On the ResNet18 model running on the low-power Arm Cortex-A53 CPU in the Raspberry Pi 3B+, our overall implementation realizes speedups of up to 2.9x on 2-bit and 4.4x on 1-bit over an optimized floating-point baseline improving upon the results published in prior works [7]. Additionally, we also extend our work to object detection models and achieve speedups of up to 2.2x and 3.2x over TFLite with XNNPACK [10][12] and ONNX Runtime [11], respectively, for both YOLOv5s and YOLOv5m [1] on the Arm Cortex-A72 CPU in the Raspberry Pi 4B [19].

## VI. End-to-end Pipeline for computer vision tasks

Our end-to-end pipeline combines all the steps required to quantize and run a CNN based model on DeepliteRT. As Figure 3 shows, Deeplite Neutrino™ automatically quantizes a trained full-precision CNN down to 2 bits and passes it to Deeplite Compiler that then compiles the quantized model and generates a *dlrt* file ready to be deployed and executed with DeepliteRT.

## VII. Evaluation

We evaluated the accuracy and inference time of ultra-low bit models for classification and object detection tasks. For classification tasks, we used ResNet18 [3] and ResNet50 [3] on ImageNet dataset and ResNet18 on VWW [8] dataset. We also successfully benchmarked our pipeline on VGG16-SSD [17], YOLOv5s [1] and YOLOv5m [1] on Pascal VOC [9] and subset of MS-COCO [20] datasets for object detection. To run these models, we have used TensorFlow [15], PyTorch [16], ONNX Runtime [11], TVM [28], TFLite [10] and DeepliteRT. The target platforms used for these experiments include the Raspberry Pi 3B+ with 4x Arm Cortex-A53, the Raspberry Pi 4B with 4x Arm Cortex-A72 and the NVIDIA Jetson Nano with 4x Arm Cortex-A57.

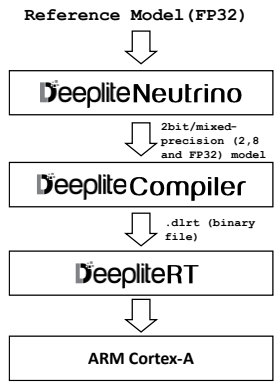

Fig. 3. End-to-end quantization and runtime pipeline

### A. ResNet18 on VWW

Figure 5 illustrates the accuracy-performance tradeoff for DeepliteRT and ONNX Runtime on the ResNet18 model trained on VWW dataset for 2A/2W (2 bits for activations and weights) and 1A/2W (1 bit for activation and 2 bits for weights) precisions. The quantized ResNet18 on VWW dataset achieved 15.58x reduction in model size with 3.75x and 2.90x speedups on Raspberry Pi 3B+ and Raspberry Pi 4B devices, respectively. Moreover, this was achieved with less than a 2% drop in accuracy for the 1A/2W configuration and less than a 1% drop in accuracy for the 2A/2W configuration.

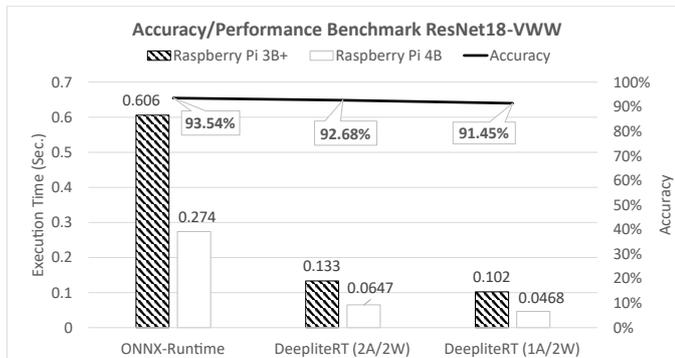

Fig. 4. Accuracy/performance benchmark of DeepliteRT on ResNet18 model on VWW dataset (2A/2W- weights and activations quantized to 2 bits, 1A/2W- activations quantized to 1 bit and weights quantized to 2 bits)

We also benchmarked DeepliteRT against TFLite [10] with XNNpack [12] for 2A/2W and 1A/2W quantized models. As Figure 6 shows, our 2-bit model has less than 1% accuracy drop while being significantly faster compared to the TFLite INT8 model running on the XNNpack backend.

### B. VGG16-SSD on VOC

We used the Single Shot Detection (SSD) [17] model with VGG16 [31] backbone as a part of our object detection performance analysis. As evident from the graph in figure 7, we achieved 3.19x and 2.95x speedup on Raspberry Pi 3B+ and Raspberry Pi 4B, respectively. The most impressive aspect of the quantized model is that the significant speedup and compression come at the cost of less than 0.02 drop in mAP.

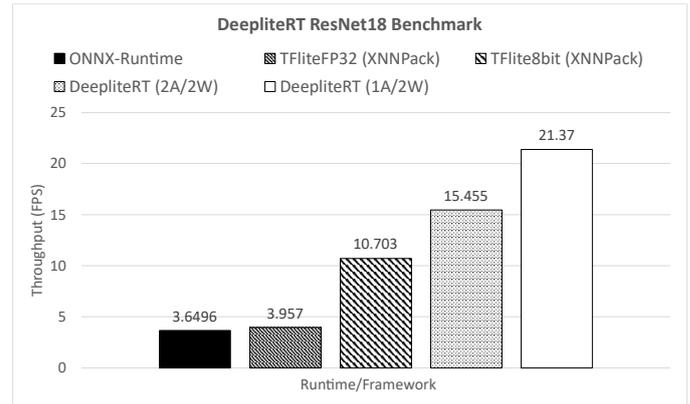

Fig. 5. DeepliteRT performance benchmark for ResNet18 with VWW [8] dataset 224px image

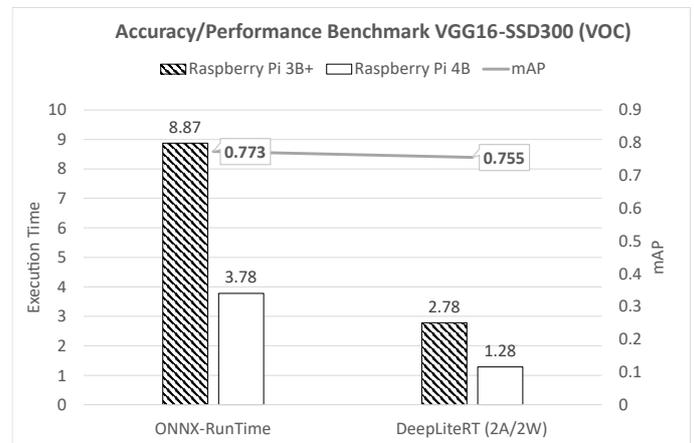

Fig. 6. Accuracy/performance benchmark of DeepliteRT on VGG16-SSD300 [17] model on Pascal VOC [9] object detection dataset (2A/2W- weights and activations quantized to 2 bits)

### C. YOLOv5s and YOLOv5m on VOC

Our classification results are state-of-the-art for both accuracy and latency on Arm hardware with limited computational power such as the Cortex A-53 in Raspberry Pi 3B+. VGG16-SSD [17] detection results were also promising but even the best performing model on the Raspberry Pi 4B takes more than a second for a single inference, so it might not be practical to utilize it for commercial applications. To address this challenge, we extended our quantization support to YOLOv5 [1], a state-of-the-art object detection architecture. To understand how our ultra-low precision models compare to TFLite with XNNPack [10] [12 (FP16 model from Ultralytics repository [1]) and full-precision ONNX Runtime results on Arm CPUs, we used YOLOv5s and YOLOv5m from the Ultralytics repository and trained them on the person class in VOC [9] dataset.

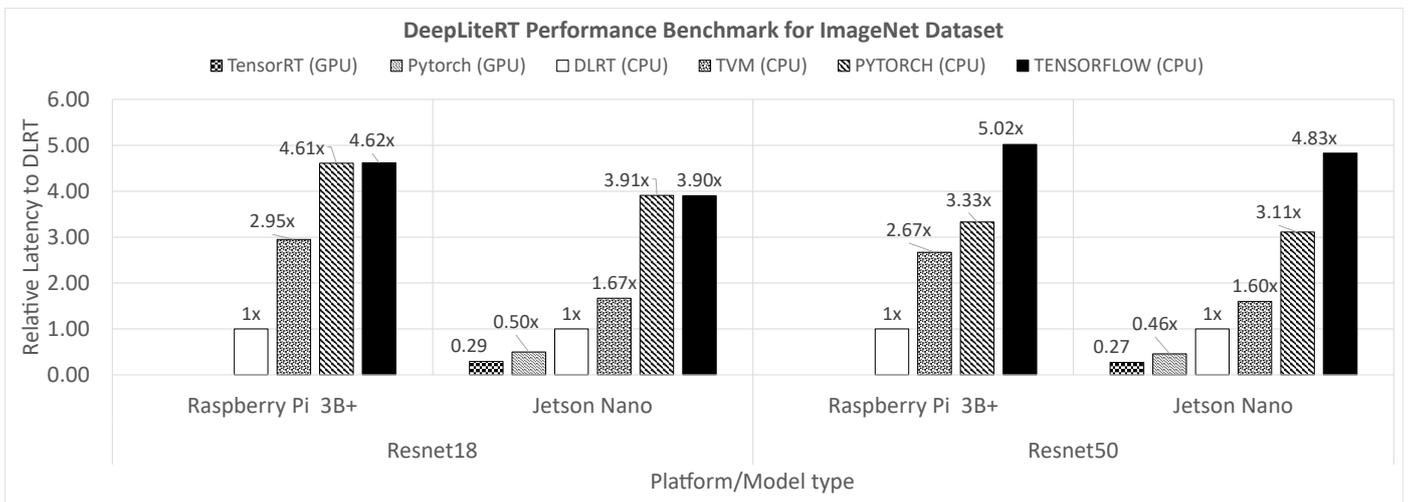

Fig. 7. Accuracy/performance benchmark of DeepliteRT on ResNet18 and ResNet50 models on ImageNet dataset, Lower bars mean faster acceleration or lower inference time, DLRT is only 50% slower compared to Embedded GPU performance

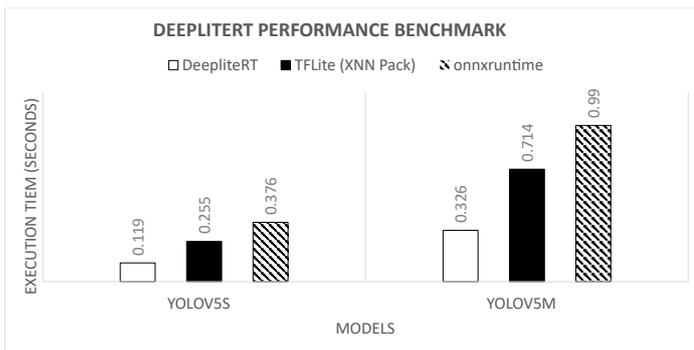

Fig. 8. DeepliteRT performance benchmark for YoloV5m and Yolo5Vs on 320px image (TFLite performs poorly without the XNNPack delegate and can be even slower compared to the FP32 models)

Our performance benchmarks show speedups of up to 2.2x over TFLite with XNNPACK [10][12] and 3.2x over ONNX Runtime [11] for both YOLOv5s and YOLOv5m [1] on Raspberry Pi 4B. As Figure 8 shows, we can even achieve 9 FPS for YOLOv5s and 3 FPS for YOLOv5m on this target device. By combining accurate, SOTA models like YOLOv5 with accessible hardware like the Arm Cortex-A CPU inside the Raspberry Pi [19], we are creating new possibilities for vision applications at the edge for the AI community.

*D. YOLOv5n on COCO 8 Classes*

We have performed an extensive evaluation of our quantization framework and runtime on a subset of MS-COCO dataset. COCO dataset consists of 80 classes, but we have evaluated it on a subset including the classes that are relevant for real life use cases. This subset includes the person, dog, cat, car, bus, truck, bicycle, and motorcycle classes. Table 1 shows the latency improvement of our 2-bit model running on Arm Cortex A-53 processor. Inference time for our quantized model is 2.54x faster compared to the baseline with approximately 1% accuracy drop. Since it is extremely challenging to quantize already compact models, we use a mixed precision approach, keeping a few quantization-sensitive layers in FP32 and the rest quantized down to 2 bits. Based on the results, mixed precision ultra-low bit quantization is an effective method to increase speedup of compact models with a minimal drop in accuracy.

TABLE I. DEEPLITERT BENCHMARK ON YOLOV5N /COCO-8 CLASSES

| Model | 352px, ARM Cortex A-53, COCO 8 classes | | |
|---|---|---|---|
| | *Quantization approach* | *mAP* | *Latency (ms)* |
| YOLOv5n -FP32 | No quantization | 0.424 | 250 |
| YOLOv5n - Mixed-precision (FP32 and 2-bit) | Conservative | 0.414 | 98.371 |

## VIII. CONCLUSION

This paper presents a novel ultra-low bit quantization method and corresponding runtime that enables executing ultra-low bit quantized models on Arm CPUs. The clear memory benefits (up to 16x compression with 2-bit quantization) are complemented by faster arithmetic enabled on low-cost CPUs. The result is 2-5x speedup over existing FP32 and INT8 runtime frameworks, approaching GPU-level latency on a commodity Arm processor as demonstrated by the image classification benchmarks on the Arm Cortex-A57 CPU in the NVIDIA Jetson Nano. We successfully applied quantization on complex SOTA models including VGG16-SSD, YOLO family and ResNet18/50. DeepliteRT yielded significant inference acceleration over next-best-alternatives, reinforcing the promise of ultra-low bit CNN models on low-cost CPUs.